\documentclass[runningheads]{llncs}
\usepackage{graphicx}
\usepackage{tikz}
\usepackage{comment}
\usepackage{amsmath,amssymb} 
\usepackage{color}

\usepackage{subfig}
\usepackage{soul}

\usepackage{xspace}
\makeatletter
\DeclareRobustCommand\onedot{\futurelet\@let@token\@onedot}
\def\@onedot{\ifx\@let@token.\else.\null\fi\xspace}
\def\eg{\emph{e.g}\onedot} 
\def\ie{\emph{i.e}\onedot}

\def\etal{\emph{et al}\onedot}
\makeatother

\usepackage{cite}
\usepackage{mathtools}
\usepackage[nice]{nicefrac}
\usepackage{todonotes}
\usepackage{bm}

\DeclareMathOperator*{\argmax}{argmax}

\begin{document}

\pagestyle{headings}
\mainmatter
\def\ECCVSubNumber{4521}  

\title{Invertible Neural BRDF\\%
for Object Inverse Rendering} 

%
\author{Zhe Chen \and Shohei Nobuhara \and Ko Nishino}
%
%
\institute{Kyoto University, Kyoto, Japan\\
\email{zchen@vision.ist.i.kyoto-u.ac.jp \{nob,kon\}@i.kyoto-u.ac.jp}\\
\url{https://vision.ist.i.kyoto-u.ac.jp}
}
\maketitle

\begin{abstract}
We introduce a novel neural network-based BRDF model and a Bayesian framework for object inverse rendering, \ie, joint estimation of reflectance and natural illumination from a single image of an object of known geometry. The BRDF is expressed with an invertible neural network, namely, normalizing flow, which provides the expressive power of a high-dimensional representation, computational simplicity of a compact analytical model, and physical plausibility of a real-world BRDF. We extract the latent space of real-world reflectance by conditioning this model, which directly results in a strong reflectance prior. We refer to this model as the invertible neural BRDF model (iBRDF). We also devise a deep illumination prior by leveraging the structural bias of deep neural networks. By integrating this novel BRDF model and reflectance and illumination priors in a MAP estimation formulation, we show that this joint estimation can be computed efficiently with stochastic gradient descent. We experimentally validate the accuracy of the invertible neural BRDF model on a large number of measured data and demonstrate its use in object inverse rendering on a number of synthetic and real images. The results show new ways in which deep neural networks can help solve challenging radiometric inverse problems.

\end{abstract}

\section{Introduction}

Disentangling the complex appearance of an object into its physical constituents, namely the reflectance, illumination, and geometry, can reveal rich semantic information about the object and its environment. The reflectance informs the material composition, the illumination reveals the surroundings, and the geometry makes explicit the object shape. Geometry recovery with strong assumptions on the other constituents has enjoyed a long history of research, culminating in various methods of shape-from-X. Accurate estimation of reflectance and illumination is equally critical for a broad range of applications, including augmented reality, robotics, and graphics where the problem is often referred to as inverse rendering. This paper particularly concerns inverse rendering of object appearance rather than scenes, where the latter would require modeling of global light transport in addition to the complex local surface light interaction.

Even when we assume that the geometry of the object is known or already estimated, the difficulty of joint estimation of the remaining reflectance and illumination, persists. The key challenge lies in the inherent ambiguity between the two, both in color and frequency \cite{raviinv}. Past methods have relied on strongly constrained representations, for instance, by employing low-dimensional parametric models, either physically-based or data-driven (\eg, Lambertian and spherical harmonics, respectively). On top of these compact parametric models, strong analytical but simplistic constraints are often required to better condition the joint estimation, such as a Gaussian-mixture on the variation of real-world reflectance and gradient and entropy priors on nonparametric illumination \cite{lombardi2016reflectance}.

While these methods based on low-dimensional parametric BRDF models have shown success in object inverse rendering ``in the wild,'' the accuracy of the estimates are inevitably limited by the expressive power of the models. As also empirically shown by Lombardi and Nishino \cite{lombardi2016reflectance}, the estimation accuracy of the two radiometric constituents are bounded by the highest frequency of either of the two. Although we are theoretically limited to this bound, low-dimensional parametric models further restrict the recoverable frequency characteristics to the approximation accuracy of the models themselves. Ideally, we would like to use high-dimensional representations for both the reflectance and illumination, so that the estimation accuracy is not bounded by their parametric forms. The challenge then becomes expressing complex real-world reflectance with a common high-dimensional representation while taming the variability of real-world reflectance and illumination so that they can be estimated from single images. Nonparametric (\ie, tabulated) BRDF representations and regular deep generative models such as generative adversarial networks \cite{goodfellow2014generative} and variational autoencoders \cite{vae} are unsuitable for the task as they do not lend a straightforward means for adequate sampling in the angular domain of the BRDF. 

In this paper, we introduce the \textit{invertible neural BRDF} model (\textit{iBRDF}) for joint estimation of reflectance and illumination from a single image of object appearance. We show that this combination of an invertible, differentiable model that has the expressive power better than a nonparametric representation together with a MAP formulation with differentiable rendering enable efficient, accurate real-world object inverse rendering. We exploit the inherent structure of the reflectance by modeling its bidirectional reflectance distribution function (BRDF) as an invertible neural network, namely, a nonlinearly transformed parametric distribution based on normalized flow \cite{Tabak_CMS10,Tabak_CPAM13,Muller:2019:NIS:3341165.3341156}. In sharp contrast to past methods that use low-dimensional parametric models, the deep generative neural network makes no assumptions on the underlying distribution and expresses the complex angular distributions of the BRDF with a series of non-linear transformations applied to a simple input distribution. We will show that this provides us with comparable or superior expressiveness to nonparametric representations. Moreover, the invertibility of the model ensures Helmholtz reciprocity and energy conservation, which are essential for physical plausibility. In addition, although we do not pursue in this paper, this invertibility makes iBRDF also suitable for forward rendering applications due to its bidirectional, differentiable bijective mapping. We also show that multiple ``lobes'' of this nonparametric BRDF model can be combined to express complex color and frequency characteristics of real-world BRDFs. Furthermore, to model the intrinsic structure of the reflectance variation of real-world materials, we condition this generative model to extract a parametric embedding space. This embedding of BRDFs in a simple parametric distribution provides us a strong prior for estimating the reflectance.

For the illumination, we employ a nonparametric representation by modeling it as a collection of point sources in the angular space (\ie, equirectangular environment map). Past methods heavily relied on simplistic assumptions that can be translated into analytical constraints to tame the high-dimensional complexity associated with this nonparametric illumination representation. Instead, we constrain the illumination to represent realistic natural environments by exploiting the structural bias induced by a deep neural network (\ie, deep image prior \cite{ulyanov2018deep}). We device this deep illumination prior by encoding the illumination as the output of an encoder-decoder deep neural network and by optimizing its parameters on a fixed random image input. 

We derive a Bayesian object inverse rendering framework by combining the deep illumination prior together with the invertible neural BRDF and a differentiable renderer to evaluate the likelihood. Due to the full differentiability of the BRDF and illumination models and priors, the estimation can be achieved through backpropagation with stochastic gradient descent. We demonstrate the effectiveness of our novel BRDF model, its embedding, deep illumination prior, and joint estimation on a large amount of synthetic and real data. Both the quantitative evaluation and quantitative validation show that they lead to accurate object inverse rendering of real-world materials taken under complex natural illumination.

To summarize, our technical contributions consist of
\begin{itemize}
    \item a novel BRDF model with the expressiveness of a nonparametric representation and computational simplicity of an  analytical distribution model,
    \item a reflectance prior based on the embedding of this novel BRDF model,
    \item an illumination prior leveraging architectural bias of neural networks,
    \item and a fully differentiable joint estimation framework for reflectance and illumination based on these novel models and priors.
\end{itemize}
We believe these contributions open new avenues of research towards fully leveraging deep neural networks in solving radiometric inverse problems.

\section{Related Work}

Reflectance modeling and radiometric quantity estimation from images has a long history of research in computer vision and related areas, studied under the umbrella of physics-based vision, appearance modeling, and inverse rendering. Here we briefly review works most relevant to ours.

\paragraph{\textbf{Reflectance Models.}}
For describing local light transport at a surface point, Nicodemus \cite{NicoBRDF} introduced the bidirectional reflectance distribution function (BRDF) as a 4D reflectance function of incident and exitant light directions. Since then, many parametric reflectance models that provide an analytical expression to this abstract function have been proposed. Empirical models like Phong \cite{phong1975illumination} and Blinn models \cite{blinn1977models} offer great simplicity for forward rendering, yet fail to capture complex reflectance properties of real-world materials making them unsuitable for reflectance estimation. Physically-based reflectance models such as Torrance-Sparrow \cite{torrance1967theory}, Cook-Torrance \cite{cook1982reflectance} and Disney material models \cite{burley2012physically} rigorously model the light interaction with micro-surface geometry. While these models capture challenging reflectance properties like off-specular reflection, their accuracy is limited to certain types of materials.

Data-driven reflectance models instead directly model the BRDF by fitting basis functions (\eg, Zernike polynomials \cite{Koenderink_ECCV96} and spherical harmonics \cite{Basri01,raviinv} or by extracting such bases from measured data \cite{matusik2003data}. Nishino \etal. introduce the directional statistics BRDF model \cite{nishino2009directional,Ko_JOSA11} based on a newly derived hemispherical exponential power distribution to express BRDFs in half-way vector representations and use their embedding as a prior for various inverse-rendering tasks \cite{lombardi2016reflectance,Oxholm_PAMI16,Lombardi_3DV16}. Ashikhmin and Premoze \cite{ashikhmin2007distribution} use a modified anisotropic Phong fit to measured data. The expressive power of these models are restricted by the underlying analytical distributions. Romeiro \etal. \cite{romeiro2008passive} instead directly use tabulated 2D isotropic reflectance distributions. Although nonparametric and expressive, using such models for estimating the reflectance remains challenging due to their high-dimensionality and lack of differentiability. We show that our invertible neural BRDF model with comparable number of parameters achieves higher accuracy while being differentiable and invertible.

\paragraph{\textbf{Reflectance Estimation.}} Joint estimation of reflectance and illumination is challenging due to the inherent ambiguity between the two radiometric quantities. For this reason, many works estimate one of the two assuming the other is known. For reflectance estimation early work such as that by Sato \etal. \cite{sato1997object} assume Torrance-Sparrow reflection and a point light source. Romeiro \etal. \cite{romeiro2008passive} estimate a nonparametric bivariate BRDF of an object of known geometry taken under known natural illumination. Rematas \cite{rematas2016deep} propose an end-to-end neural network to estimate the reflectance map. The geometry is first estimated from the image after which a sparse reflectance map is reconstructed. A convolutional neural network (ConvNet) was learned to fill the holes of this sparse reflectance map. The illumination is, however, baked into and inseparable from this reflectance map. Meka \etal. \cite{meka2018lime} models the reflectance as a linear combination of Lambertian, specular albedo and shininess and regress each component with deep convolutional auto-encoders. Lombardi and Nishino \cite{lombardi2012single} use a learned prior for natural materials using the DSBRDF model \cite{nishino2009directional} for multi-material estimation. Kang \etal. \cite{kang2018efficient} and Gao \etal. \cite{gao2019deep} estimate spatially-varying BRDFs of planar surfaces taken under designed lighting patterns with an auto-encoder. These methods are limited to known or prescribed illumination and cannot be applied to images taken under arbitrary natural illumination.

\paragraph{\textbf{Illumination Estimation.}} Assuming Lambertian surfaces, Marschner \etal. \cite{marschner1997inverse} recover the illumination using image-based bases. Garon \etal. \cite{Garon_2019_CVPR} encode the illumination with 5th-order spherical harmonics and regress their coefficients with a ConvNet. Gardner \etal. \cite{Gardner_2019_ICCV} represent lighting as a set of discrete 3D lights with geometric and photometric parameters and regress their coefficients with a neural network. LeGendre \etal. \cite{legendre2019deeplight} trained a ConvNet to directly regress a high-dynamic range illumination map from an low-dynamic range image. While there are many more recent works on illumination estimation, jointly estimating the reflectance adds another level of complexity due to the intricate surface reflection that cannot be disentangled with such methods. 

\paragraph{\textbf{Joint Estimation of Reflectance and Illumination.}} Romeiro \etal. \cite{romeiro2010blind} use non-negative matrix factorization to extract reflectance bases and Haar wavelets to represent illumination to estimate both. Their method, however, is restricted to monochromatic reflectance estimation and cannot handle the complex interplay of illumination and reflectance across different color channels. Lombardi and Nishino \cite{Lombardi_ECCV12,lombardi2016reflectance} introduce a maximum a posterior estimation framework using the DSBRDF model and its embedding space as a reflectance prior and gradient and entropy priors on nonparametric illumination. Our Bayesian framework follows their formulation but overcomes the limitations induced by the rigid parametric reflectance model and artificial priors on the illumination that leads to oversmoothing. More recently, Georgoulis \etal. \cite{georgoulis2017reflectance} extend their prior work \cite{rematas2016deep} to jointly estimate geometry, material and illumination. The method, however, assumes Phong BRDF which significantly restricts its applicability to real-world materials. Wang \etal. \cite{wang2018joint} leverage image sets of objects of different materials taken under the same illumination to jointly estimate material and illumination. This requirement is unrealistic for general inverse-rendering. Yu \etal. \cite{yu19inverserendernet} introduce a deep neural network-based inverse rendering of outdoor imagery. The method, however, fundamentally relies on Lambertian reflectance and low-frequency illumination expressed in spherical harmonics. Azinovi\'{c} \etal. \cite{azinovic2019inverse} introduce a differentiable Monte Carlo renderer for inverse path tracing indoor scenes. Their method employs a parametric reflectance model \cite{burley2012physically}, which restricts the types of materials it can handle. While the overall differentiable estimation framework resembles ours, our work focuses on high fidelity inverse rendering of object appearance by fully leveraging a novel BRDF model and deep priors on both the reflectance and illumination. We believe our method can be integrated with such methods as \cite{azinovic2019inverse} for scene inverse rendering in the future.

\section{Differentiable Bayesian Joint Estimation}

We first introduce the overall joint estimation framework. Following \cite{lombardi2016reflectance}, we formulate object inverse rendering, namely joint estimation of reflectance and illumination of an object of known geometry, from a single image as maximum a posteriori (MAP) estimation
\begin{equation}
    \argmax_{\mathbf{R}, \mathbf{L}} p(\mathbf{R}, \mathbf{L} | \mathbf{I}) \propto p(\mathbf{I} | \mathbf{R}, \mathbf{L})p(\mathbf{R})p(\mathbf{L})\,,
    \label{eq:map}
\end{equation}
where $\mathbf{I}$ is the RGB image, $\mathbf{R}$ is the reflectance, and $\mathbf{L}$ is the environmental illumination. We assume that the camera is calibrated both geometrically and radiometrically, and the illumination is directional (\ie, it can be represented with an environment map). Our key contributions lie in devising an expressive high-dimensional yet invertible model for the reflectance $\mathbf{R}$, and employing strong realistic priors on the reflectance $p(\mathbf{R})$ and illumination $p(\mathbf{L})$.

We model the likelihood with a Laplacian distribution on the log radiance for robust estimation \cite{lombardi2016reflectance,wilcox2011introduction}
\begin{equation}
    p(\mathbf{I} | \mathbf{R}, \mathbf{L}) = \prod_{i, c}\frac{1}{2b_I}\exp{\left[-\frac{\left|\log I_{i, c} - \log E_{i, c}(\mathbf{R}, \mathbf{L})\right|}{b_I}\right]}\,,
\end{equation}
where $b_I$ controls the scale of the distribution, $I_{i, c}$ is the irradiance at pixel $i$ in color channel $c$, and $E_{i, c}(\mathbf{R}, \mathbf{L})$ is the expectation of the rendered radiance. 

To evaluate the likelihood, we need access to the forward rendered radiance $E_{i,c}(\mathbf{R},\mathbf{L})$ as well as their derivatives with respect to the reflectance and illumination, $\mathbf{R}$ and $\mathbf{L}$, respectively. For this, we implement the differential path tracing method introduced by Lombardi and Nishino \cite{Lombardi_3DV16}. 

\section{Invertible Neural BRDF}

Real-world materials exhibit a large variation in their reflectance that are hard to capture with generic bases such as Zernike polynomials or analytic distributions like that in the Cook-Torrance model. Nonparametric representations, such as simple 3D tabulation of an isotropic BRDF, would better capture the wide variety while ensuring accurate representations for individual BRDFs. On the other hand, we also need to differentiate the error function Eq. \ref{eq:map} with respect to the reflectance and also evaluate the exact likelihood of the BRDF. For this, past methods favored low-dimensional parametric reflectance models, which have limited expressive power. We resolve this fundamental dilemma by introducing a high-dimensional parametric reflectance model based on an invertible deep generative neural network. To be precise, the BRDF is expressed as a 3D reflectance distribution which is transformed from a simple parametric distribution (\eg, uniform distribution). The key property is that this transformation is a cascade of invertible linear transforms that collectively results in a complex nonparametric distribution. The resulting parameterization is high-dimensional having as many parameters as a nonparametric tabulation, thus being extremely expressive, yet differentiable and also can be efficiently sampled and evaluated.

\subsection{Preliminaries}

The BRDF defines point reflectance as the ratio of reflected radiance to incident irradiance given the 2D incident and exitant directions, $\mathbf{\omega}_i$ and $\mathbf{\omega}_o$, respectively, 
\begin{equation}
    f_r(\mathbf{\omega}_i, \mathbf{\omega}_o) = \frac{\text{d}L_r(\mathbf{\omega}_o)}{L_i(\mathbf{\omega}_i)\cos{\theta_i}\text{d}\mathbf{\omega}_i}\,,
\end{equation}
where $\omega_i = (\theta_i. \phi_i)$, $\omega_o = (\theta_o. \phi_o)$, $\theta \in [0, \frac{\pi}{2}]$, $\phi \in [0, 2\pi]$. 

We use the halfway-vector representation introduced by Rusinkiewicz \cite{Rusinkiewicz}
\begin{align}
    \mathbf{\omega}_h = \frac{\mathbf{\omega}_i + \mathbf{\omega}_o}{\left\|\mathbf{\omega}_i + \mathbf{\omega}_o\right\|}\,,\,\, 
    \mathbf{\omega}_d = \mathsf{R}_\mathbf{b}(-\theta_h)\mathsf{R}_\mathbf{n}(-\phi_h)\mathbf{\omega}_i\,,
\end{align}
where $\mathsf{R}_\mathbf{a}$ denotes a rotation matrix around a 3D vector $\mathbf{a}$, and $\mathbf{b}$ and $\mathbf{n}$ are the binormal and normal vectors, respectively. The BRDF is then a function of the difference vector and the halfway vector $f_r(\theta_h, \phi_h, \theta_d, \phi_d)$. We assume isotropic BRDFs, for which $\phi_h$ can be dropped: $f_r(\theta_h, \theta_d, \phi_d)$. Note that $f_r$ is a 3D function that returns a 3D vector of RGB radiance when given a unit irradiance.

\subsection{BRDF As An Invertible Neural Network}

Our goal in representing the reflectance of real-world materials is to derive an expressive BRDF model particularly suitable for estimating it from an image. Given that deep neural networks lend us complex yet differentiable functions, a naive approach would be to express the (isotropic) BRDF with a neural network. Such a simplistic approach will, however, break down for a number of reasons. 

For one, materials with similar reflectance properties (\ie, kurtosis of angular distribution) but with different magnitudes will have to be represented independently. This limitation can be overcome by leveraging the fact that generally each color channel of the same material shares similar reflectance properties. We encode the BRDF as the product of a normalized base distribution $p_r$ and a color vector $\mathbf{c} = \{r, g, b\}$
\begin{equation}
    f_r(\theta_h, \theta_d, \phi_d) = p_r(\theta_h, \theta_d, \phi_d) \mathbf{c}\,.\label{eq:f_r}
\end{equation}
The color vector is left as a latent parameter and the base distribution is fit with density estimation during training as we explain later. This way, we can represent materials that only differ in reflectance intensity and base colors but have the same distribution properties by sharing the same base distribution. This separation of base color and distribution also lets us model each BRDF with superpositions of $f_r$, \ie, employ multiple ``lobes.'' This is particularly important when estimating reflectance, as real-world materials usually require at least two colored distributions due to their neutral interface reflection property (\eg, diffuse and illumination colors) \cite{Lee:1992:ICS:136809.136907}.

The bigger problem with simply using an arbitrary deep neural network to represent the BRDF is their lack of an efficient means to sample (e.g., GANs) and restricting latent parametric distribution (e.g., VAEs). For this, we turn to normalizing flow models \cite{Tabak_CMS10,Tabak_CPAM13} which is one of a family of so-called invertible neural networks \cite{Ardizzone_2019}.
\begin{figure}[t]
    \centering
    \subfloat[][]{\includegraphics[width=0.55\linewidth]{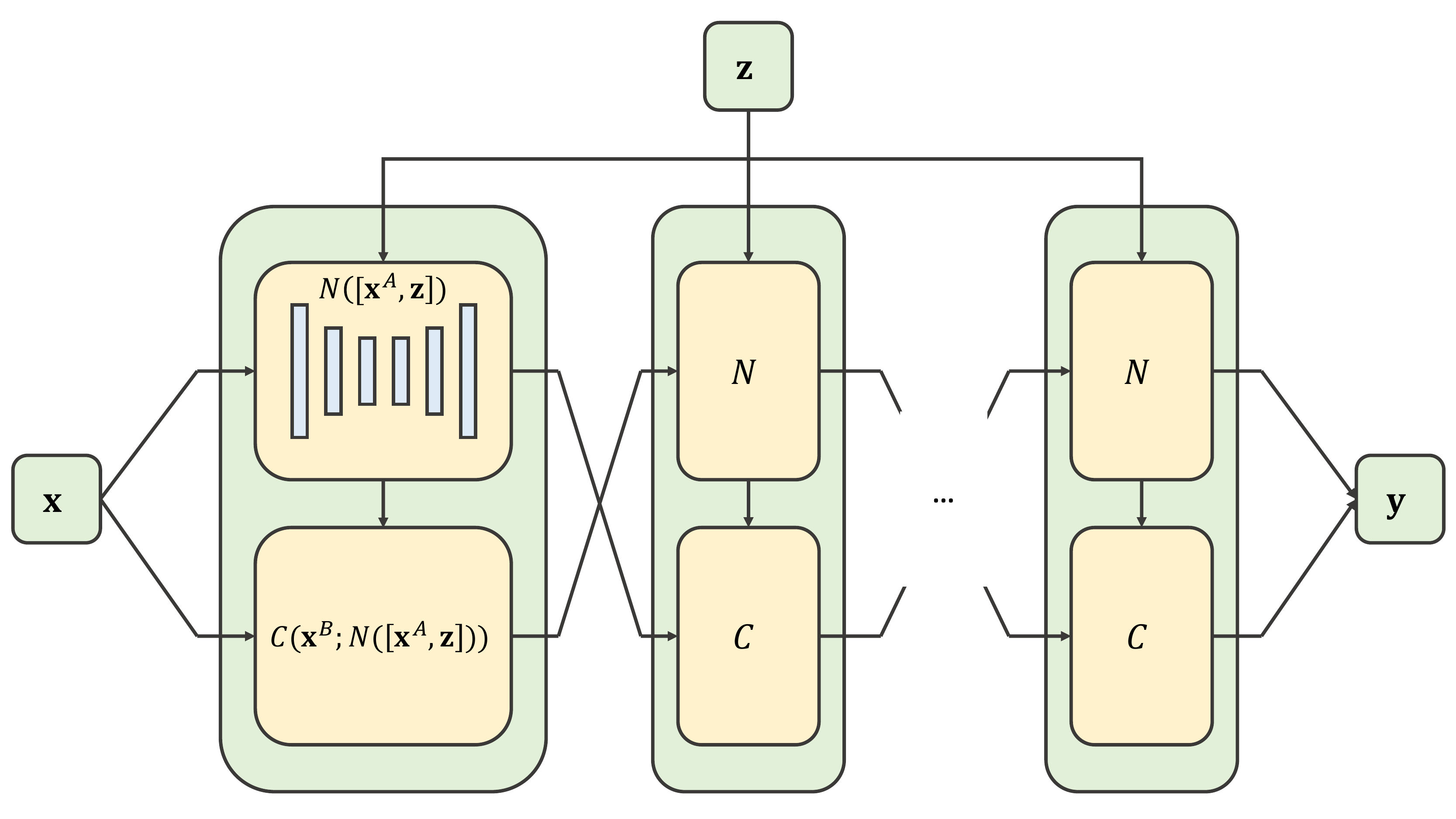}}~~
    \subfloat[][]{\includegraphics[width=0.3\linewidth]{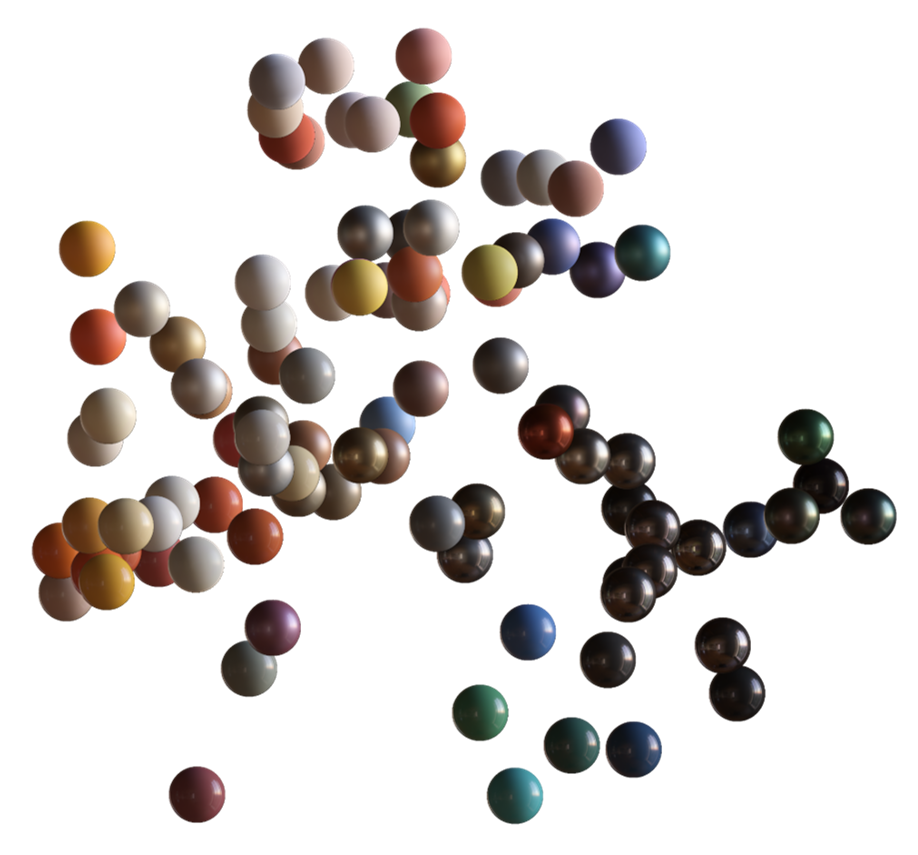}}
    \caption{(a) Architecture of the invertible neural BRDF model. The base distribution ($p_r$ in Eq. \ref{eq:f_r}) is represented with a normalizing flow \cite{Dinh_2014b,Muller:2019:NIS:3341165.3341156}, which transforms the input 3D uniform distribution $q(\mathbf{x})$ into a BRDF $p_r(\mathbf{y})$ through a cascade of bijective transformations. (b) We condition the parameters with a code $\mathbf{z}$, which let's us learn an embedding space of real-world BRDFs (shown in 2D with PCA). Similar materials are grouped together and arranged in an intuitive manner in this continuous latent space which can directly be used as a reflectance prior.}
    \label{fig:brdf-arch-latent}
\end{figure}

In particular, we represent the normalized base distribution using an extended Non-linear Independent Components Estimation (NICE) model \cite{Dinh_2014b,Muller:2019:NIS:3341165.3341156}. Given a tractable latent (\ie, input) distribution $q(\mathbf{x})$ and a sequence of bijective transformations $f = f_1 \circ f_2 \circ \dots \circ f_N$, we can get a new distribution $p_r(\mathbf{y})$ by applying transformations $\mathbf{y} = f(\mathbf{x}; \mathbf{\Theta})$. Since $f$ is bijective, under the change of variable formula, $q(\mathbf{x})$ and $p(\mathbf{y})$ are linked with
\begin{equation}
    p_r(\mathbf{y}) = q(\mathbf{x})\left|\det\left(\frac{\text{d}\mathbf{x}}{\text{d}\mathbf{y}}\right)\right| = q(f^{-1}(\mathbf{y};\mathbf{\Theta}))\left|\det\left( \frac{\text{d}f^{-1}(\mathbf{y};\mathbf{\Theta})}{\text{d}\mathbf{y}} \right)\right|\,,
\end{equation}
where $\left|\det(\frac{\text{d}f^{-1}(\mathbf{y}; \mathbf{\Theta})}{\text{d}\mathbf{y}})\right|$ is the absolute value of the Jacobian determinant. Such a sequence of invertible transformations $f$ is called a normalizing flow. As long as $f$ is complex enough, we can get an arbitrarily complicated $p_r(\mathbf{y})$ in theory. As shown in Fig. \ref{fig:brdf-arch-latent}(a), for each layer $f$, we use the coupling transformation family $C$ proposed in NICE and each layer is parameterized with a UNet $N$. The input $\mathbf{x}$ is split into two parts $\mathbf{x}^A$ and $\mathbf{x}^B$, where $\mathbf{x}^A$ is left unchanged and fed into $N$ to produce the parameters of the transformation $C$ that is applied to $\mathbf{x}^B$. Then $\mathbf{x}^A$ and $C(\mathbf{x}^B; N(\mathbf{x}^A))$ are concatenated to give the output. For the input latent distribution $q(\mathbf{x})$, we use a simple 3D uniform distribution. 

There are some practical concerns to address before we can train this invertible neural BRDF on measured data. Since the base distribution $p_r(\theta_h, \theta_d, \phi_d)$  of a BRDF inherently has a finite domain ($\theta_h \in [0, \frac{\pi}{2})$, $\theta_d \in [0, \frac{\pi}{2})$, $\phi_d \in [0, \pi)$), it will be easier to learn a transformation mapping from a tractable distribution with the same domain to it. Thus we make $q(\theta_h, \theta_d, \phi_d) \sim \mathcal{U}^3(0, 1)$ and normalize each dimension of $p_r(\theta_h, \theta_d, \phi_d)$ to be in $[0, 1)$ before training. Then we adopt the piecewise-quadratic coupling transformation \cite{Muller:2019:NIS:3341165.3341156} as $C$ to ensure that the transformed distribution has the same domain. This corresponds to a set of learnable monotonically increasing quadratic curve mappings from $[0, 1)$ to $[0, 1)$ each for one dimension.

\subsection{Latent Space of Reflectance}

Expressing the rich variability of real-world BRDFs concisely but with a differentiable expression is required to devise a strong constraint (\ie, prior) on the otherwise ill-posed estimation problem (Eq. \ref{eq:map}). As a general approach, such a prior can be derived by embedding the BRDFs, for instance, the 100 measured BRDFs in the MERL database \cite{matusik2003data}, in a low-dimensional parametric space. Past related works have achieved this by modeling the latent space in the parameter space of the BRDF model (\eg, linear subspace in the DSBRDF parameter space \cite{lombardi2016reflectance}). More recent works \cite{gao2019deep,kang2018efficient} train an auto-encoder on the spatial map of analytical BRDF models to model spatially-varying BRDF. The latent space together with the decoder of a trained auto-encoder is then used for material estimation. Our focus is instead on extracting a tight latent space various real-world materials span in the nonparametric space of invertible neural BRDFs.

We achieve this by conditioning the invertible neural BRDF on a latent vector $\mathbf{z}$ (Fig. \ref{fig:brdf-arch-latent}(a)) . We refer to this vector as the embedding code to avoid confusion with the latent distribution (\ie, input) to the invertible BRDF. We jointly learn the parameters $\mathbf{\Theta}$ of iBRDF and its embedding code $\mathbf{z}$ 
\begin{equation}
    \argmax_{\mathbf{z}, \mathbf{\Theta}}  \frac{1}{M}\sum_i^M  \frac{1}{N}\sum_j^N \log p_r(\theta_h^{ij}, \theta_d^{ij}, \phi_d^{ij} | \mathbf{z}^i; \mathbf{\Theta})\,.
\end{equation}
Each trained embedding code in the embedding space is associated with a material in the training data (\eg, one measured BRDF of the MERL database). This formulation is similar to the generative latent optimization \cite{bojanowski2017optimizing} where the embedding code is directly optimized without the need of an encoder.

We treat each color channel of the training measured data as an independent distribution. Each distribution is assigned a separate embedding code which is initialized with a unit Gaussian distribution. Additionally, after each training step, we project $\mathbf{z}$ back into the unit hypersphere to encourage learning a compact latent space. This is analogous to the bottleneck structure in an auto-encoder. This conditional invertible neural BRDF is trained by maximizing the likelihood.

During inference, we fix $\mathbf{\Theta}$ and optimize the embedding code $\mathbf{z}$ to estimate the BRDF. In practice, we set the dimension of $\mathbf{z}$ to 16. In other words, the invertible neural BRDFs of real-world materials are embedding in a 16D linear subspace. Figure \ref{fig:brdf-arch-latent}(b) shows the embedding of the 100 measured BRDFs in the MERL database after training. Materials with similar properties such as glossiness lie near each other forming an intuitive and physically-plausible latent space that can directly be used as a strong reflectance prior. Since all materials are constrained within the unit hypersphere during training, we thus define $p(\mathbf{R}) \sim \mathcal{N}(0, \sigma^2 I)$.

\section{Deep Illumination Prior}

Incident illumination to the object surface can be represented in various forms. With the same reason that motivated the derivation of iBRDF, we should avoid unnecessary approximation errors inherent to models and represent the illumination as a nonparametric distribution. This can be achieved by simply using a latitude-longitude panoramic HDR image $\mathbf{L}(\theta, \phi)$ as an environment map to represent the illumination \cite{Debevec:2008:RSO:1401132.1401175}, which we refer to as the illumination map. 

The expressive power of a nonparametric illumination representation, however, comes with increased complexity. Even a $1^\circ$--sampled RGB illumination map ($360\times 180$ in pixels) would have $194400$ parameters in total. To make matters worse, gradients computed in the optimization are sparse and noisy and would not sufficiently constrain such a high degree of freedom. Past work have mitigated this problem by imposing strong analytical priors on the illumination $p(\mathbf{L})$, such as sparse gradients and low entropy \cite{lombardi2016reflectance} so that they themselves are differentiable. These analytical priors, however, do not necessarily capture the properties of natural environmental illumination and often lead to overly simplified illumination estimates. 

We instead directly constrain the illumination maps to be ``natural'' by leveraging the structural bias of deep neural networks. Ulyanov \etal. \cite{ulyanov2018deep} found that the structure of a deep neural network has the characteristic of impeding noise and, in turn, represent natural images without the need of pre-training. That is, an untrained ConvNet can directly be used as a natural image prior. We adopt this idea of a deep image prior and design a deep neural network for use as a deep illumination prior.

We redefine the illumination map as $\mathbf{L}=g(\theta, \phi; \mathbf{\Phi})$, where $\mathbf{\Phi}$ denotes the parameters of a deep neural network $g$. We estimate the illumination map as an ``image'' generated by the deep neural network $g$ with a fixed random noise image as input by optimizing the network parameters $\mathbf{\Phi}$. The posterior (Eq. \ref{eq:map}) becomes
\begin{equation}
    \argmax_{\mathbf{R}, \mathbf{\Phi}} p(\mathbf{R}, \mathbf{L} | \mathbf{I}) \propto p(\mathbf{I} | \mathbf{R}, g(\theta, \phi; \mathbf{\Phi}))p(\mathbf{R})\,.
\end{equation}

The advantages of this deep illumination prior is twofold. First, the structure of a deep neural network provides impedance to noise introduced by the differentiable renderer and stochastic gradient descent through the optimization. Second, angular samples of the illumination map are now interdependent through the network, which mitigates the problem of sparse gradients. For the deep neural network $g$, we use a modified UNet architecture \cite{unet}. To avoid the checkerboard artifacts brought by transposed convolutional layer, we use a convolutional layer followed by a bilinear upsampling for each upsampling step. Additionally, to preserve finer details, skip connections are added to earlier layers.

\begin{figure}[t]
    \centering
    \includegraphics[width=1\linewidth]{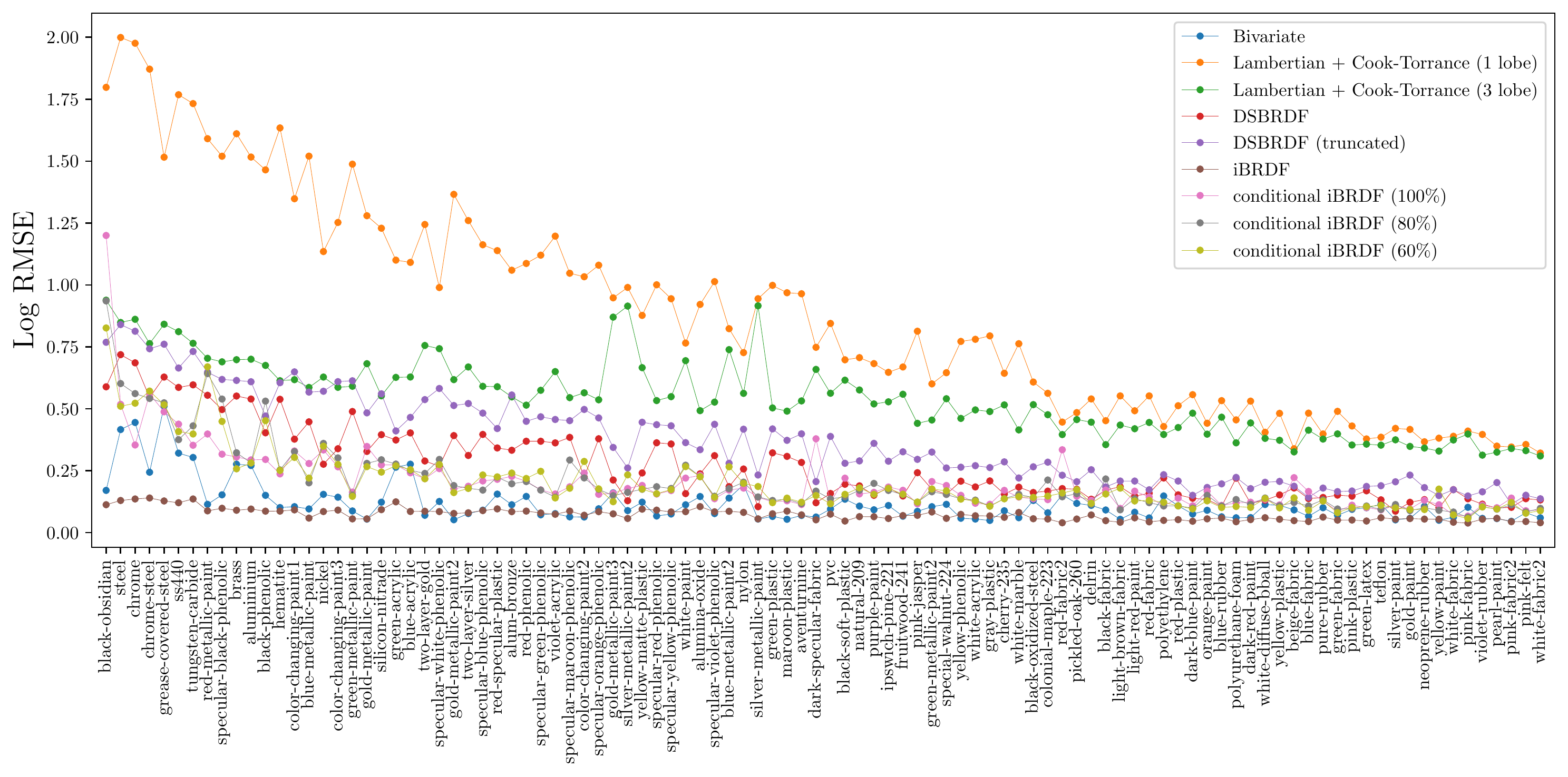}
    \caption{Log RMSE of iBRDF and conditional iBRDF (\ie, iBRDF with learned latent space) for 100 MERL materials. The iBRDF has higher accuracy than a nonparametric bivariate model. The conditional iBRDF models using 100\%, 80\%, and 60\% of the leave-one-out training data achieves higher accuracy than other parametric models (\ie, DSBRDF and Cook-Torrance). These results show the expressive power and generalizability of the invertible neural BRDF model.}
    \label{fig:fit-brdf}
\end{figure}

\section{Experimental Results}

We conducted a number of experiments to validate the effectiveness of 1) the invertible neural BRDF model for representing reflectance of real-world materials; 2) the conditional invertible neural BRDF model for BRDF estimation; 3) the deep illumination prior, and 3) the Bayesian estimation framework integrated with the model and priors for single-image inverse rendering. 

\subsection{Accuracy of Invertible Neural BRDF} \label{sec:accuracy}

To evaluate the accuracy of invertible neural BRDF, we learn its parameters to express measured BRDF data in the MERL database and evaluate the representation accuracy using the root mean squared error (RMSE) in log space\cite{lombardi2016reflectance}.

As Fig. \ref{fig:fit-brdf} shows, the invertible neural BRDF achieves higher accuracy than the nonparametric bivariate BRDF model. The conditional iBRDF, learned on 100\%, 80\%, and 60\% training data all achieve high accuracy superior to other parametric models namely the DSBRDF and Cook-Torrance models. Note that all these conditional iBRDFs were trained without the test BRDF data. This resilience to varying amounts of training data demonstrates the robustness of the invertible neural BRDF model and its generalization power encoded in the learnt embedding codes. The results show that the model learns a latent space that can be used as a reflectance prior without sacrificing its expressive power.

\subsection{Reflectance Estimation with iBRDF}
\begin{figure}[t]
    \centering
    \subfloat[][]{\includegraphics[width=0.50\linewidth]{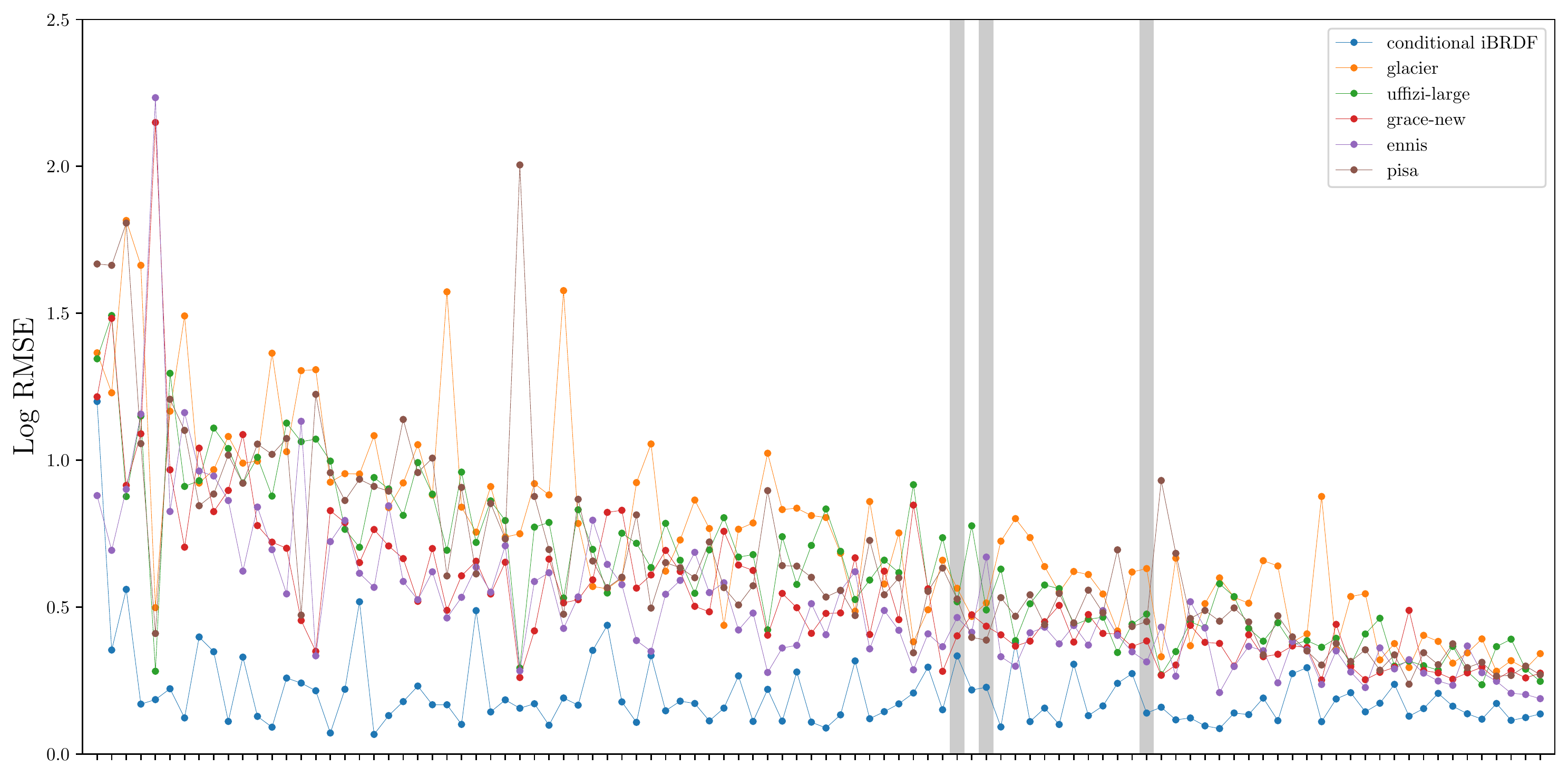}}
    \subfloat[][]{\includegraphics[width=0.47\linewidth]{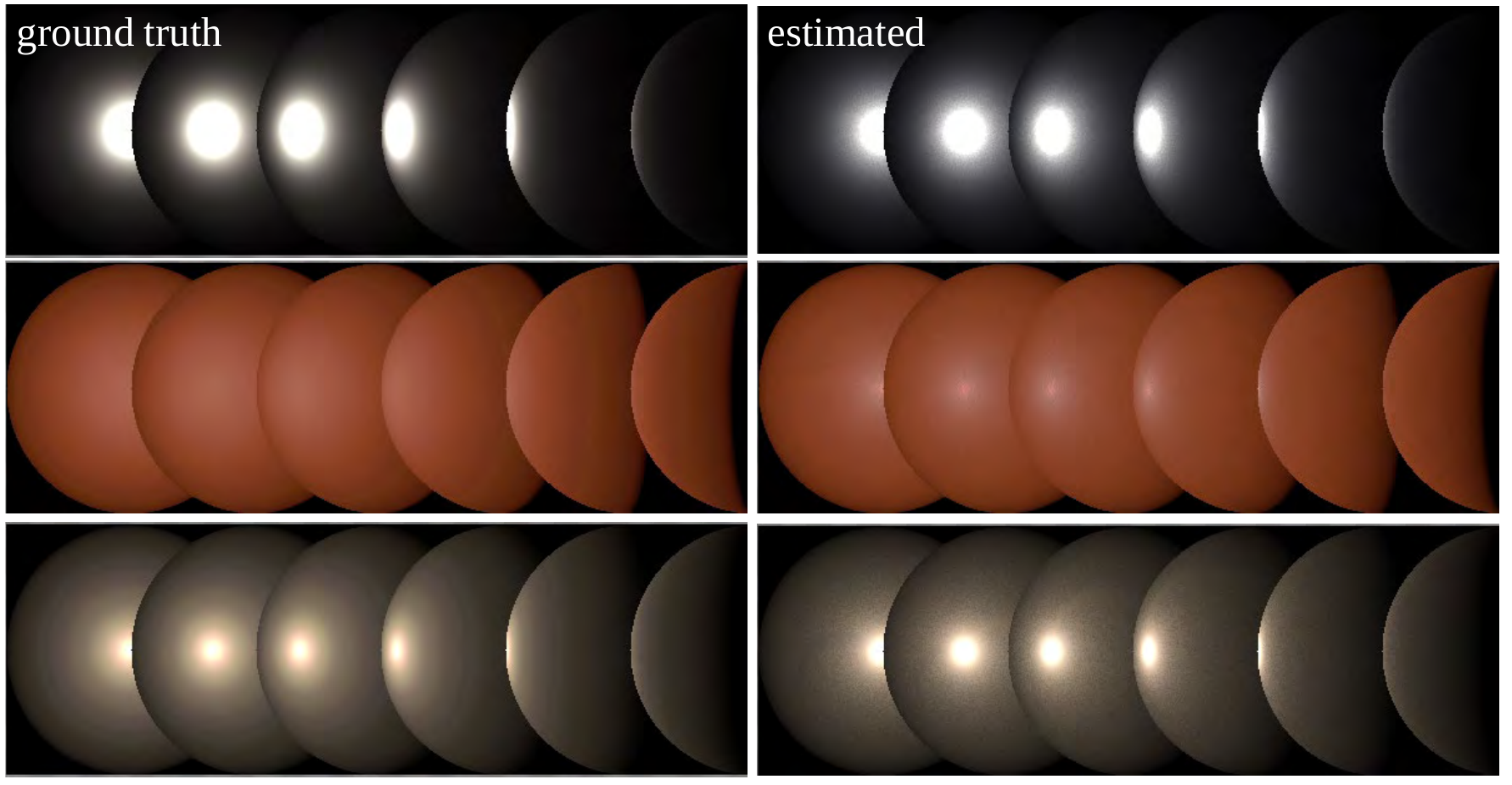}}
    \caption{(a) Log RMSE of iBRDF estimation from single images of a sphere with different materials under different known natural illumination. (b) Renderings of three samples of ground truth and estimated BRDFs. These results clearly demonstrate the high accuracy iBRDF can provide when estimating the full BRDF from partial angular observations under complex illumination.}
    \label{fig:estimate-brdf-natural-illumination}
\end{figure}
Next, we evaluate the effectiveness of the invertible neural BRDF for single-image BRDF estimation. Unlike directly fitting to measured BRDF data, the BRDF is only partially observed in the input image. Even worse, the differentiable path tracer inevitably adds noise to the estimation process. The results of these experiments tell us how well the invertible neural BRDF can extrapolate unseen slices of the reflectance function given noisy supervision, which is important for reducing ambiguity between reflectance and illumination in their joint estimation. 

We evaluate the accuracy of BRDF estimation for each of the 100 different materials in the MERL database rendered under 5 different known natural illuminations each taken from Debevec's HDR environment map set \cite{DebevecLP}. The BRDF for each material was represented with the conditional iBRDF model trained on all the measured data expect the one to be estimated (\ie, 100\% conditional iBRDF in Sec. \ref{sec:accuracy}). Fig. \ref{fig:estimate-brdf-natural-illumination}(a) shows the log-space RMS error for each of the combinations of BRDF and natural illumination. The results show that the BRDF can be estimated accurately regardless of the surrounding illumination. Fig. \ref{fig:estimate-brdf-natural-illumination}(b) shows spheres rendered with different point source directions using the recovered BRDF. The results match the ground truth measured BRDF well, demonstrating the ability of iBRDF to robustly recover the full reflectance from partial angular observations in the input image.

\subsection{Illumination Estimation with Deep Illumination Prior}

\begin{figure}[t]
    \centering
    \includegraphics[width=1\linewidth]{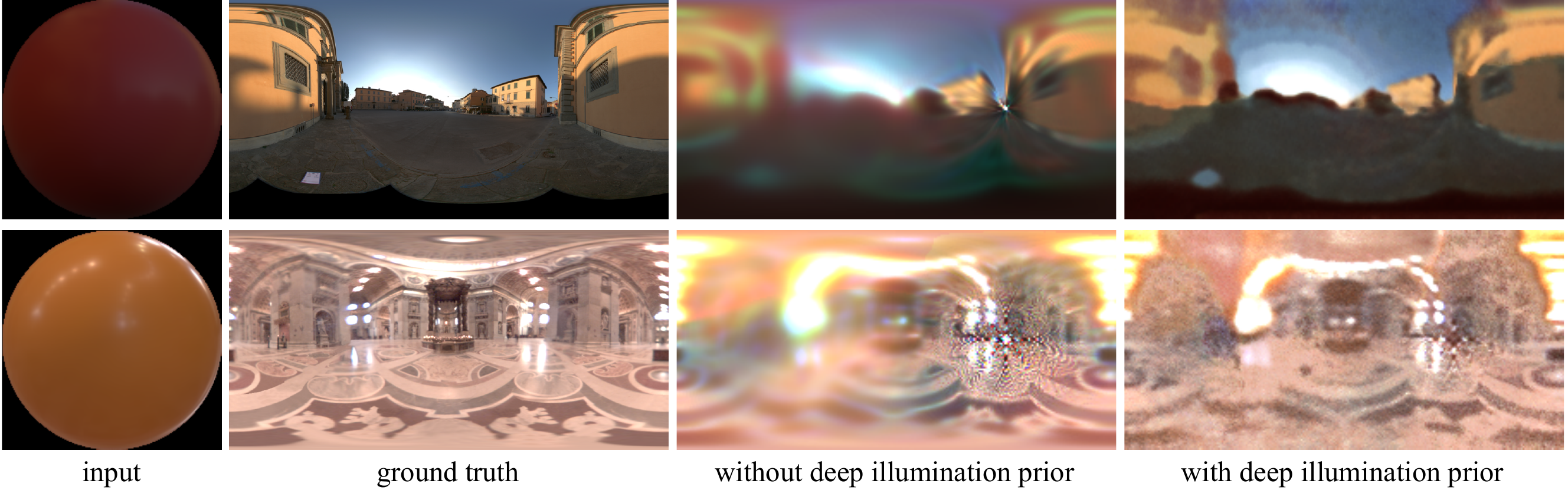}
    \caption{Results of illumination estimation without and with the deep illumination prior. The illumination estimates using the prior clearly shows higher details that match those of the ground truth.}
    \label{fig:illumination-estimation}
\end{figure}

We examine the effectiveness of the deep illumination prior by evaluating the accuracy of illumination estimation with known reflectance and geometry. We rendered images of spheres with 5 different BRDFs sampled from the MERL database under 10 different natural illuminations. Fig. \ref{fig:illumination-estimation} shows samples of the ground truth and estimated illumination without and with the deep illumination. These results clearly demonstrate the effectiveness of the proposed deep illumination prior which lends strong constraints for tackling joint estimation.

\subsection{Joint Estimation of Reflectance and Illumination}

\begin{figure}[t]
    \centering
    \subfloat[][]{\includegraphics[width=\linewidth]{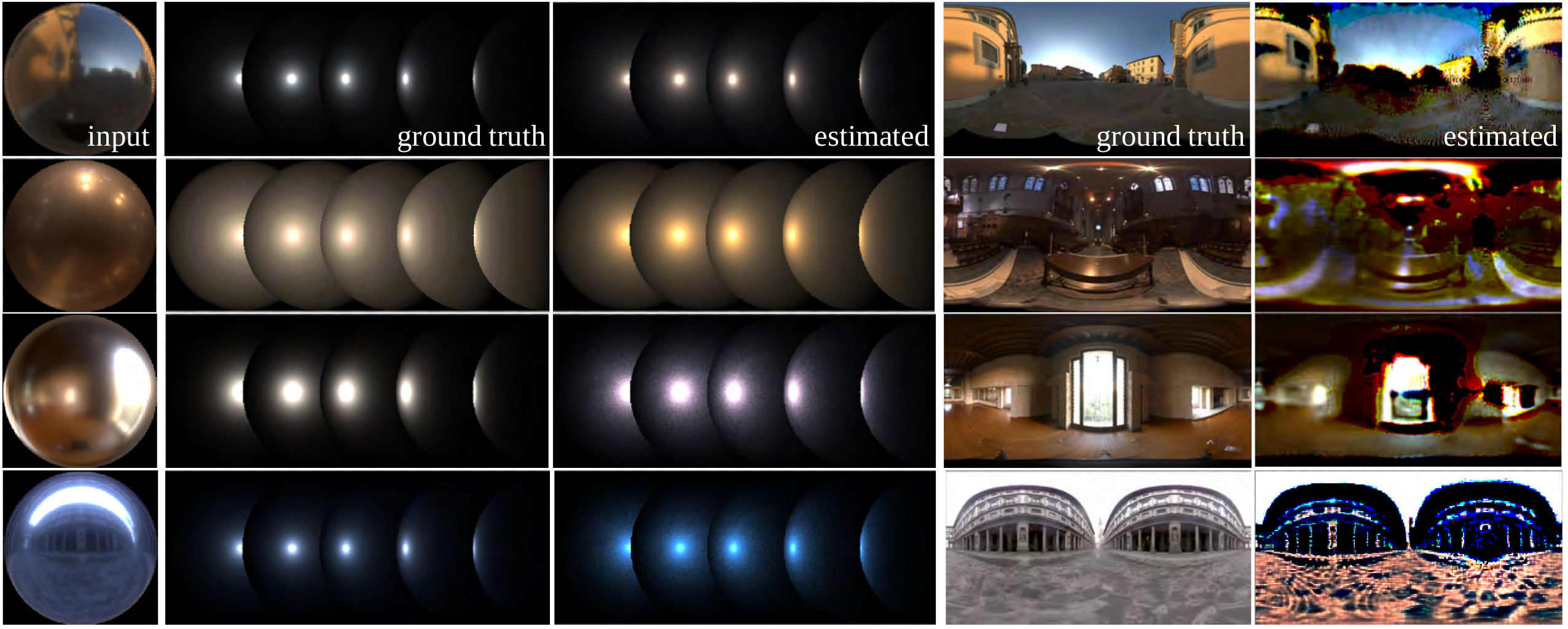}}\\
    \subfloat[][]{\includegraphics[width=\linewidth]{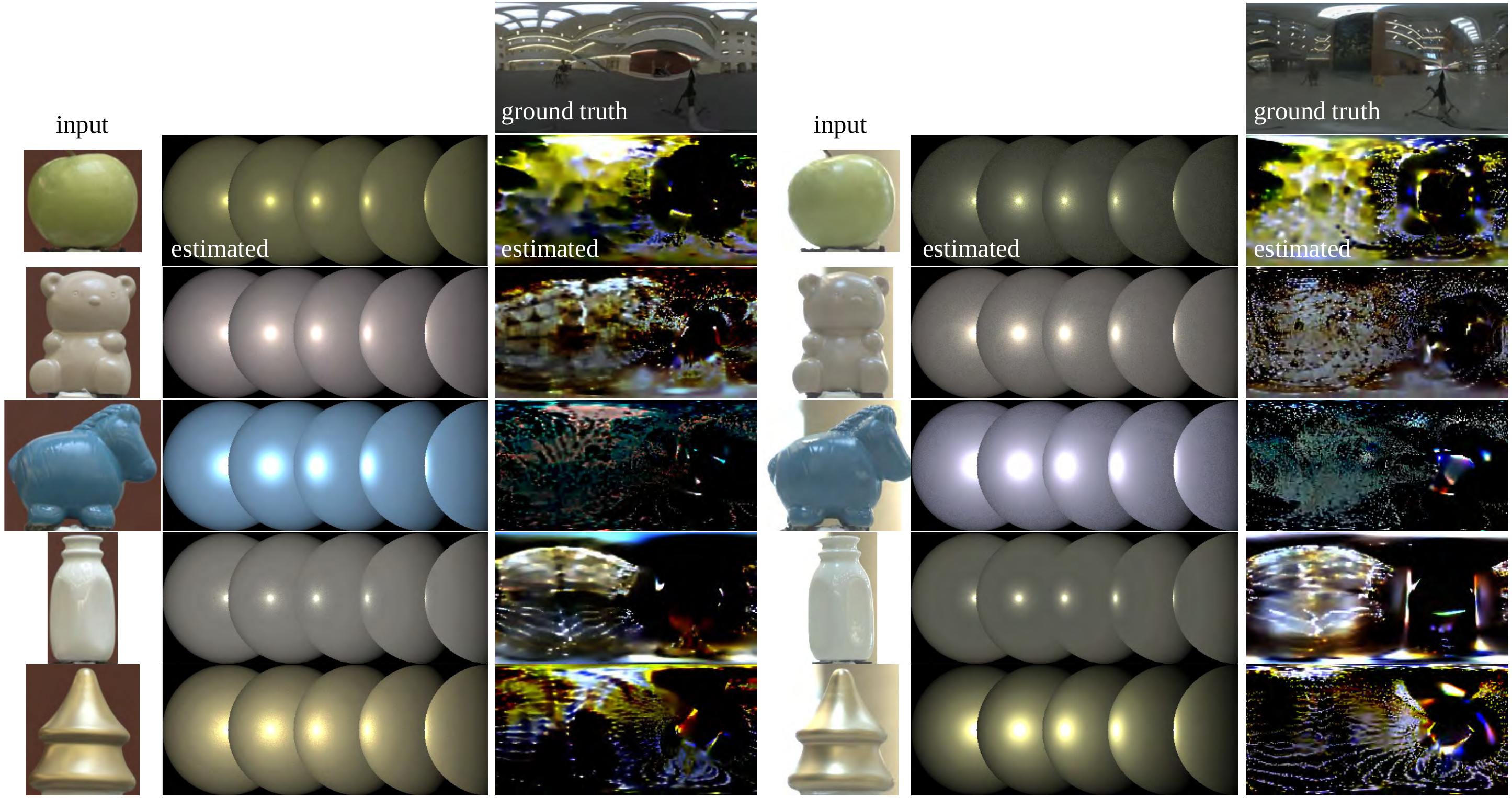}}
    \caption{Results of object inverse rendering using iBRDF and its latent space, and the deep illumination prior from (a) synthetic images and (b) real images. All results are in HDR shown with fixed exposure, which exaggerates subtle differences (\eg, floor tile pattern in Uffizi). The results show that the model successfully disentangles the complex interaction of reflectance and illumination and recovers details unattainable in past methods (\eg, \cite{lombardi2016reflectance}).}
    \label{fig:fit-real-error}
\end{figure}

We integrate iBRDF and its latent space as a reflectance prior together with the deep illumination prior into the MAP estimation framework (Eq. \ref{eq:map}) for object inverse rendering and systematically evaluate their effectiveness as a whole with synthetic and real images. First, we synthesized a total of 100 images of spheres rendered with 20 different measured BRDFs sampled from the MERL database under 5 different environment maps. Fig. \ref{fig:fit-real-error}(a) shows some of the estimation results. Qualitatively, the recovered BRDF and illumination match the ground truth well, demonstrating the effectiveness of iBRDF and priors for object inverse rendering. As evident in the illumination estimates, our method is able to recover high-frequency details that are not attainable in past methods. Please compare these results with, for instance, Fig. 7 of \cite{lombardi2016reflectance}.

Finally, we apply our method to images of real objects taken under natural illumination. We use the Objects Under Natural Illumination Database \cite{lombardi2016reflectance}. Fig \ref{fig:fit-real-error}(b) shows the results of jointly estimating the BRDF and illumination. Our reflectance estimates are more faithful to the object appearance than those by Lombardi and Nishino, and the illumination estimates have more details (compare Fig. \ref{fig:fit-real-error} with Fig. 10 of \cite{lombardi2016reflectance}), which collectively shows that our method more robustly disentangles the two from the object appearance. Note that the color shifts in the BRDF estimates arise from inherent color constancy, and the geometry dictates the recoverable portions of the environment. The estimates are in HDR and exposures are manually set to match as there is an ambiguity in global scaling. Please see the supplemental material for more results.

\section{Conclusion}

In this paper, we introduced the invertible neural BRDF model and an object inverse rendering framework that exploits its latent space as a reflectance prior and a novel deep illumination prior. Through extensive experiments on BRDF fitting, recovery, illumination estimation, and inverse rendering, we demonstrated the effectiveness of the model for representing real-world reflectance as well as its use, together with the novel priors, for jointly estimating reflectance and illumination from single images. We believe these results show new ways in which powerful deep neural networks can be leveraged in solving challenging radiometric inverse and forward problems.

\paragraph*{Acknowledgement} This work was in part supported by JSPS KAKENHI 17K20143 and a donation by HiSilicon.

\clearpage
%
%
\bibliographystyle{splncs04}
\bibliography{egbib,kon,RSSI,PbVcolor}
\end{document}


\pagestyle{headings}
\mainmatter
\def\ECCVSubNumber{4521}  

\title{Invertible Neural BRDF\\%
for Object Inverse Rendering\\
\vspace{12pt}
\underline{\large{Supplementary Material}}} 

\titlerunning{Invertible Neural BRDF for Object Inverse Rendering}
%
\author{Zhe Chen \and Shohei Nobuhara \and Ko Nishino}
%
%
\institute{Kyoto University, Kyoto, Japan\\
\email{zchen@vision.ist.i.kyoto-u.ac.jp \{nob,kon\}@i.kyoto-u.ac.jp}\\
\url{https://vision.ist.i.kyoto-u.ac.jp}
}
\maketitle

\begin{abstract}
In this supplementary material, we show additional experimental results on
\begin{itemize}
    \item in-depth comparison with Georgoulis et al. TPAMI 2017,
    \item reflectance estimation with iBRDF,
    \item illumination estimation with deep illumination prior, and
    \item joint estimatation of reflectance and illumination both for synthetic and real images.

\end{itemize}
\end{abstract}

\section{Comparison with Georgoulis et al. TPAMI 2017 [15]}

As we stated in the main manuscript, ``Georgoulis et al. [15] extend their prior work [36] to jointly estimate geometry, material and illumination. The method, however, assumes Phong BRDF which significantly restricts its applicability to real-world materials." For this reason, their data are selectively of shiny material. Furthermore, the method consists of two steps in which the first step predicts a reflectance map from the input image and the second step decomposes the reflectance map into Phong parameters and an environment map. As our goal is fundamentally different, in that we jointly estimate an arbitrary BRDF and natural illumination directly from the input image albeit of an object with known geometry, our method is compared with the reflectance map decomposition network in [15]. 

\begin{figure}[t]
    \centering
    \includegraphics[width=\linewidth]{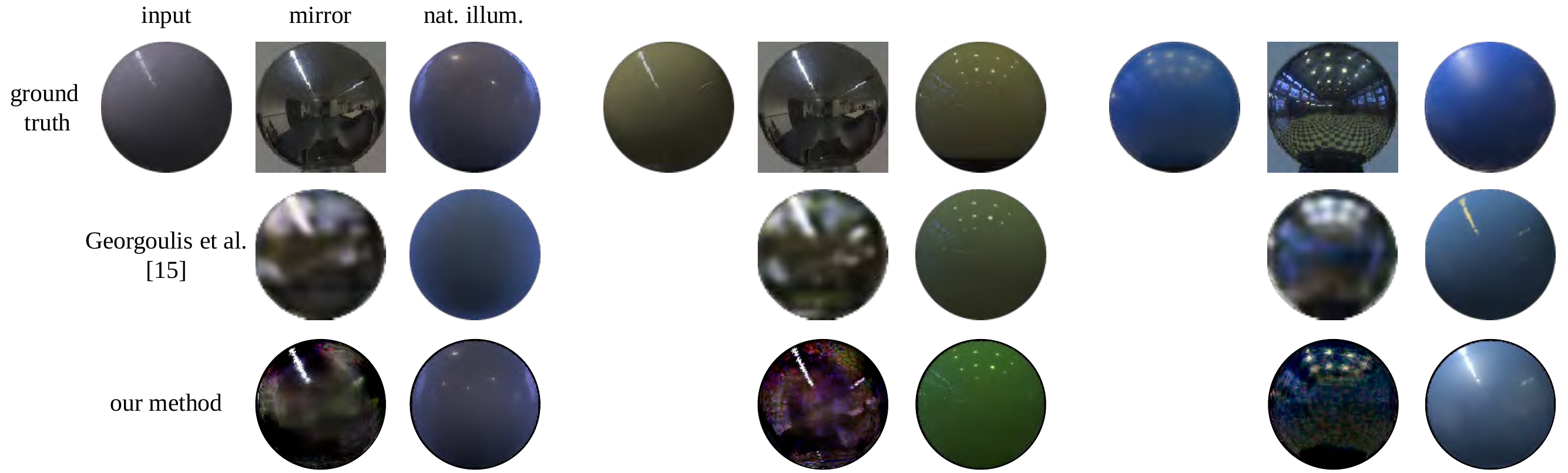}
    \caption{Comparison with DeLightNet [15]. The illumination estimate is shown as a rendered mirror sphere (mirror) and the reflectance estimate is shown as a sphere rendered under a different natural illumination (nat. illum.) from the input image (input). Our results capture finer details of the BRDF and the illumination demonstrating more robust and accurate decoupling of the two from object appearance.}
    \label{fig:delightnet_comp}
\end{figure}

Since we thoroughly evaluate our method's effectiveness on synthetic data in the main manuscript and in the following sections, we focus on comparing our method to the reflectance map decomposition on the real data of [15]. Fig. \ref{fig:delightnet_comp} shows comparisons of the estimated reflectance and illumination side-by-side with their results. We show results of rendered sphere of estimated illumination with mirror reflection (mirror) and rendered sphere of the estimated BRDF with a different illumination (nat. illum.). We omit comparison on rendered spheres of the estimated illumination with different known BRDF from the input as we believe they are misleading. As Lombardi and Nishino [25] showed, the reflectance and illumination are estimated up to the highest frequencies of either. As such, relighting using the illumination estimate with another BRDF would not accurately capture the true accuracy of the illumination estimate as that BRDF will attenuate the illumination esimate's frequency properties. Furthermore, neither the paper nor the code mentions which BRDFs were used to render these relit spheres, which prevents us from making a comparison.

Overall, judging from the sphere renderings of the estimated BRDF with a different illumination, our BRDF estimates qualitatively appear more accurate and faithful to the underlying reflectance of the input image as well as ground truth (e.g., higher frequency details of illumination estimates). Our method is a physically-based reconstruction, that decouples the reflectance and illumination of object appearance. In contrast, the method of [15] is a learned decomposition on tens of thousands of images, fundamentally bound by the combinations seen in the training data. Our method does not involve any learning, other than the BRDF model itself. We believe these two methods complement each other and can be used in conjunction, perhaps to obtain a quick learning-based initialization and then a physically-based decoupling for complex surfaces and environments that are rarely accurately represented with the Phong model. Table \ref{tab:delightnet_comp_err} shows quantitative comparison of 91 combinations of those we could identify (on the project web page--it is not clear what the remaining 9 are) among the 100 in [15]. For direct comparison to [15], we calculate the metrics in the log space defined by $\log{(x+1.0)}$ as the original paper rather than $\log{(x)}$ as used in other parts of our paper. Note that there is a fundamental ambiguity in the scale between the recovered illumination and BRDF in addition to the color. The network in [15] recovers the parameters of an analytical reflectance model (i.e., Phong), not the radiance distribution of the BRDF, which implicitly avoids this scale ambiguity. In our case, it is hard to determine the scale difference. Instead, we multiply the rendering by a scale factor that minimizes the MSE between the recovery and the ground truth. Note that this post-processing does not affect the properties of the recovered BRDF and illumination. As such the comparison is fair. As shown in Table \ref{tab:delightnet_comp_err}, we achieve lower errors in all metrics, especially for the log RMSE, even without scale correction. These results show that our estimation matches the characteristics of the ground truth very well.

\begin{table}[t]
    \centering
    \begin{tabular}{lrr|rr}
    \hline
    & \multicolumn{2}{c}{Mirror} & \multicolumn{2}{c}{Nat. Illum.}\\\hline
      method & Log RMSE & DSSIM & Log RMSE & DSSIM  \\\hline
      Gerogoulis et al. [15] & 0.933 & \textbf{0.365} & 1.110 &  0.186 \\\hline
      ours (w/ scale correction) & \textbf{0.744} & 0.369& \textbf{0.340} & \textbf{0.179} \\
      ours (w/o scale correction) & 0.926 & 0.369& 0.932 & 0.179 \\
      \hline
    \end{tabular}

    \caption{Mean Log RMSE and mean DSSIM errors of illumination estimates (mirror) and reflectance estimates (nat. illum.). Our method also quantitatively outperforms that of [15].}
    \label{tab:delightnet_comp_err}
\end{table}

\section{Reflectance Estimation with iBRDF}

\begin{figure}[t]
    \centering
    \includegraphics[width=\linewidth]{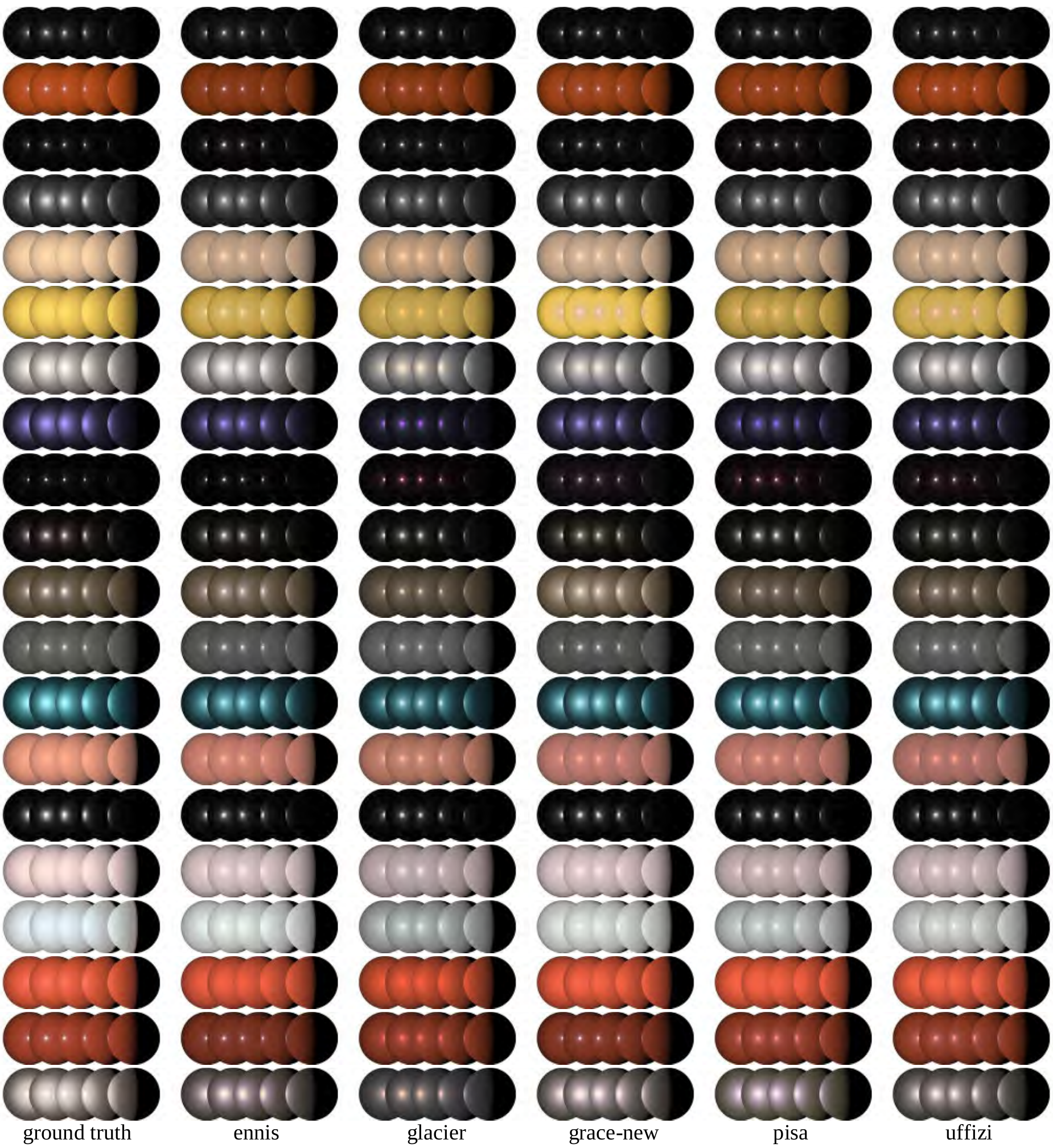}
    \caption{Additional results of reflectance estimation from a single image with known illumination. The estimated reflectance match the ground truth (left most) well for all different illuminations (right five). }
    \label{fig:brdf_full}
\end{figure}

In the main manuscript, we validated the effectiveness of the invertibile neural BRDF for single-image BRDF estimation by showing the log RMSE errors of the accuracy of BRDF estimation for each of the 100 different materials in the MERL database rendered under 5 different known natural illuminations, i.e., total 500 tests (Fig. 3(a) of main manuscript). The results show that the BRDF can be estimated accurately regardless of the surrounding illumination. Here, in Fig. \ref{fig:brdf_full}, in addition to Fig. 3(b) of the main manuscript, we show additional estimation results as spheres rendered with different point source directions using the recovered BRDF put side-by-side with the ground truth. The recovered BRDF renderings match the ground truth measured BRDF well, even when the illumination differs, demonstrating the ability of iBRDF to robustly recover the full reflectance from partial angular observations in the input image. 

\section{Illumination Estimation with Deep Illumination Prior}

\begin{figure}[t]
    \centering
    \includegraphics[width=\linewidth]{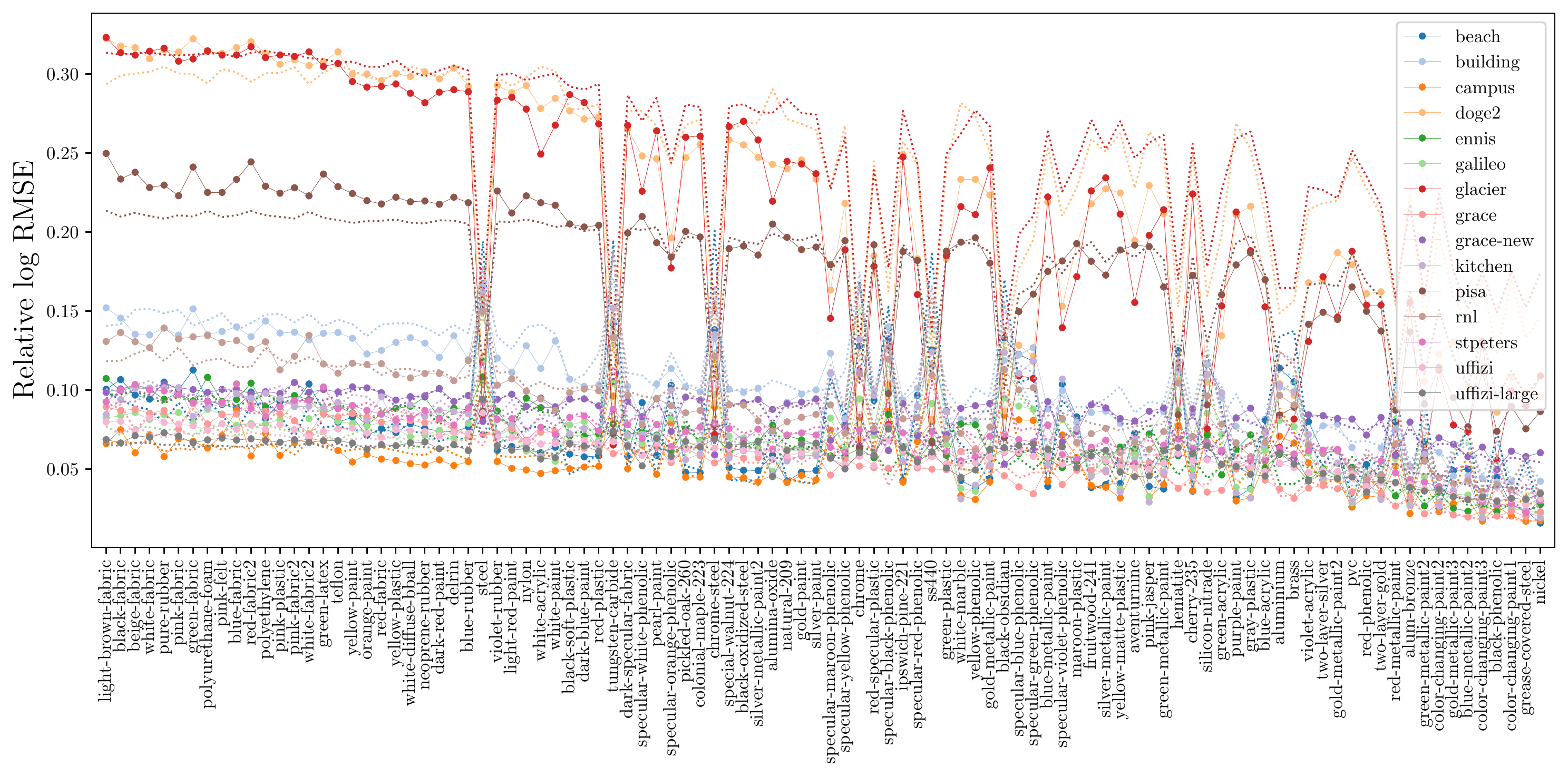}
    \caption{Relative log-space RMSE errors of estimated illumination with (solid curve with circles) and without (dashed curve) deep illumination prior for 15 different natural illumination recovered from 100 different BRDFs. For most combinations of illumination and material, the RMSE when estimated with the deep illumination prior is lower, demonstrating the effectiveness of the prior.}
    \label{fig:ill_full}
\end{figure}
We validated the effectiveness of the deep illumination prior by showing samples of the ground truth and estimated illumination without and with the deep illumination prior in the main manuscript (Fig. 4). Here we show additional quantitative analysis. We rendered images of spheres with all the 100 different BRDFs of the MERL database under 15 different natural illuminations captured as HDR environment maps. Fig. \ref{fig:ill_full} shows the relative log RMSE of the estimated illumination with (solid cirles) and without (dashed) the deep illumination prior for all 1500 combinations. We use relative log RMSE, i.e., the log RMSE normalized by the difference between the brightest and the dimmest point in the illumination, since HDR environment maps have different dynamic ranges. For all combinations, except for some involving matte materials, which results in smooth rather than structurally clear (see Fig. 4 of main manuscript) estimated illumination that RMSE favors, as well as a handful of other combinations in the total of 1500, the deep illumination prior achieves higher accuracy of illumination estimates. Table \ref{tab:ill_meanrmse} shows the mean relative log RMSE errors of the estimated illumination with and without the deep illumination prior. For all illumination, on average across different BRDFs, the deep illumination prior was effective in achieving more accurate illumination estimation. These results show that the deep illumination prior effectively constraints the optimization to recover accurate, dense non-parametric representations of a wide variety of complex, natural illumination.

\begin{table}[b]
    \centering
    \begin{tabular}{|l|r|r|r|r|r|r|r|}
    \hline
      environment map & beach & building & campus & doge2 & ennis & galileo & glacier  \\\hline
      without prior & 0.077 & 0.110 & 0.058 &  0.243 & \textbf{0.066} & 0.071 & 0.252\\\hline
      with prior & \textbf{0.073} & \textbf{0.101} & \textbf{0.053} & \textbf{0.216}& 0.068 & \textbf{0.065} & \textbf{0.207} \\
      \hline
    \end{tabular}
   \begin{tabular}{|l|r|r|r|r|r|r|r|r|}
      \hline
     environment map &  grace & grace new & kitchen & pisa & rnl & st peters & uffizi & uffizi large \\\hline
      without prior &  0.059 & \textbf{0.082} & 0.084 & 0.177 & 0.097 & 0.076 & 0.062 & 0.060\\\hline
      with prior &  \textbf{0.057} & 0.083 & \textbf{0.074} & \textbf{0.174} & \textbf{0.089} & \textbf{0.069} & \textbf{0.059} & \textbf{0.058}\\  
      \hline

    \end{tabular}
    \caption{Mean relative log-space RMSE errors of estimated illumination without and with the deep illumination prior. For every illumination, the use of the prior leads to lower error.}
    \label{tab:ill_meanrmse}
\end{table}

\section{Joint Estimation of Reflectance and Illumination}

\begin{figure}[t]
    \centering
    \subfloat[][Log-space RMSE of estimated reflectance]{\includegraphics[width=\linewidth]{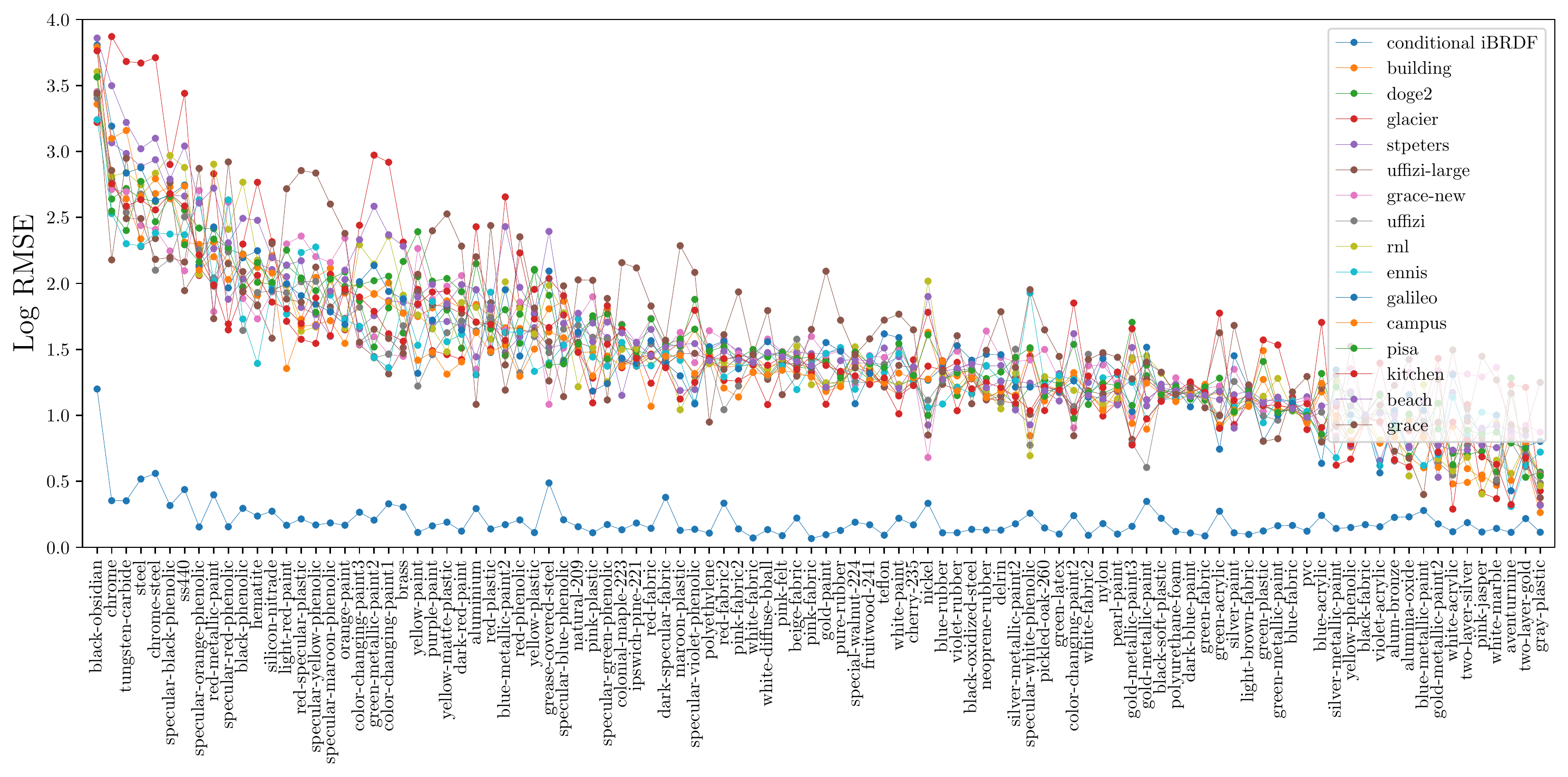}}\\
    \subfloat[][Relative log-space RMSE of estimated illumination]{\includegraphics[width=\linewidth]{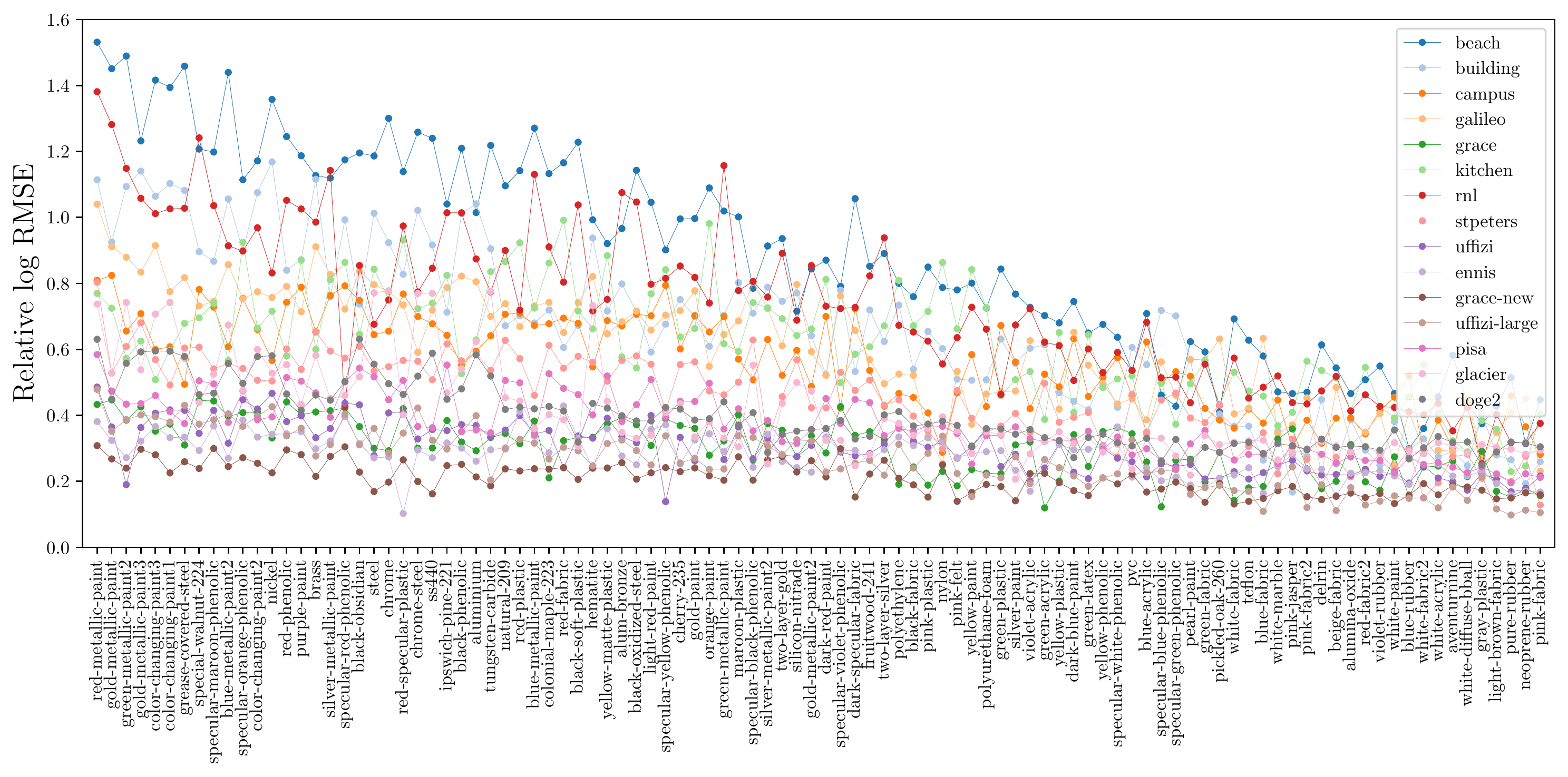}}
    \caption{Log-space RMSE of jointly estimated reflectance and illumination using the conditional invertible neural BRDF and deep illumination prior for 1500 different combinations of 100 MERL BRDFs and 15 environmentmaps. The blue curve in (a) is the BRDF fit of conditional iBRDF for reference. These extensive results demonstrate the effectiveness of our proposed models and method (see main text).}
    \label{fig:joint_synth} 
\end{figure}

We demonstrated the effectiveness of the proposed inverse rendering framework using the invertible neural BRDF, its embedding space, and the deep illumination prior on synthetic input images in Fig. 5(a) of the main manuscript. Here we show additional results including quantitative analysis. We rendered a total of 1500 images of spheres rendered with the 100 MERL BRDFs under 15 different environment maps, and used each as an input to our inverse rendering method. Fig. \ref{fig:joint_synth}(a) shows the log-space RMSE of estimated BRDF, and Fig. \ref{fig:joint_synth}(b) shows the relative log-space RMSE of estimated illumination for all the 1500 combinations of reflectance and illumination. The BRDF estimates are particularly accurate for most materials (about 90\%) considering the fact that widely used ``Lambertian + Cook-Torrance'' reflectance model approaches log RMSE of $2$ (Fig. 2 of main manuscript). The errors of illumination estimates also stay within reasonable range from the illumination estimation errors when the BRDF is known (Fig. \ref{fig:ill_full}. Note that the input, BRDF, and illumination estimates are all in high dynamic range, and small discrepancies in bright highlights can cause large RMSE errors. These results demonstrate the robustness of our inverse rendering method, the expressiveness of our invertible neurla BRDF model, and effectiveness of the deep illumination prior.

\begin{figure}[t]
    \centering
    \includegraphics[width=\linewidth]{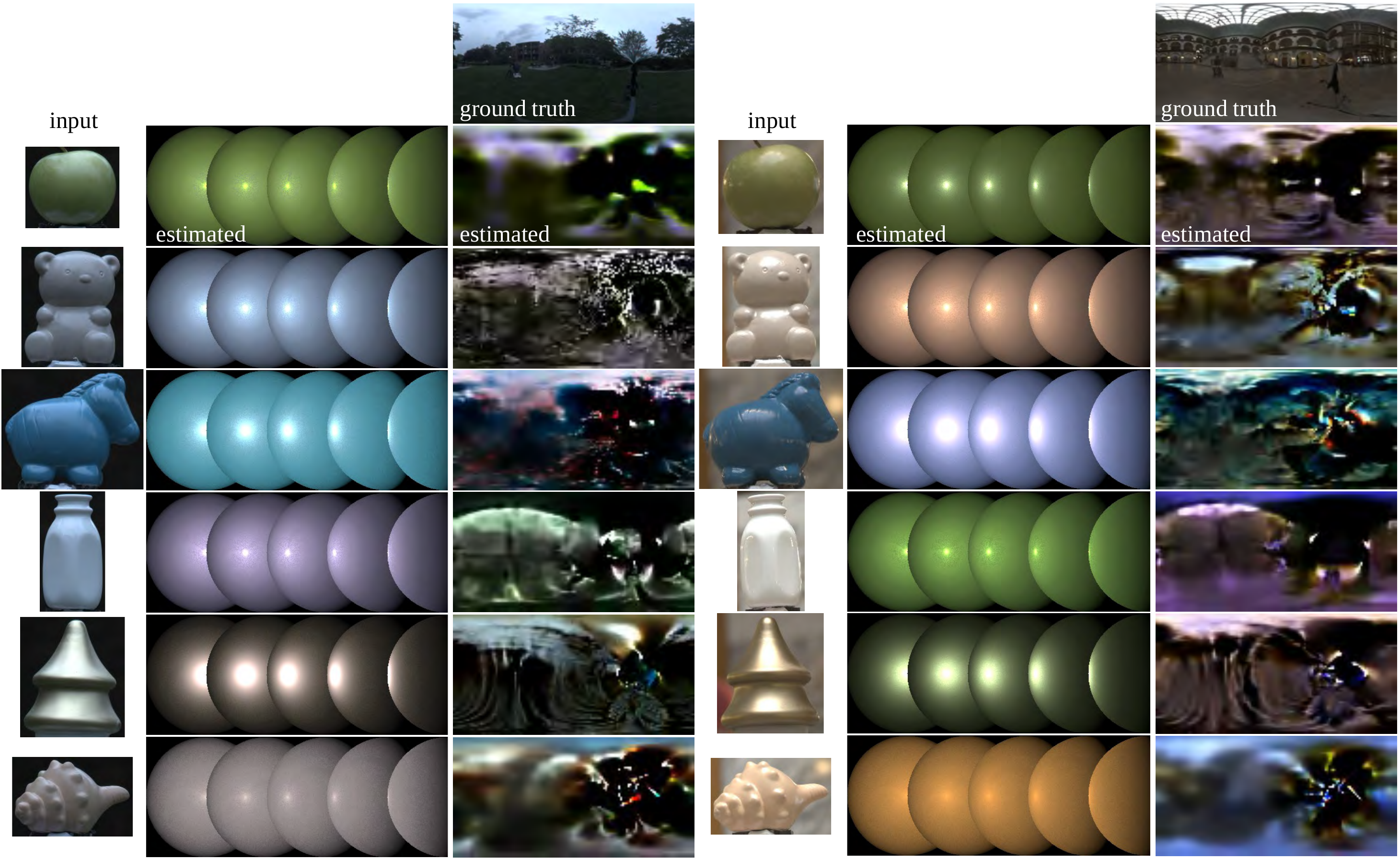}
    \includegraphics[width=0.5\linewidth]{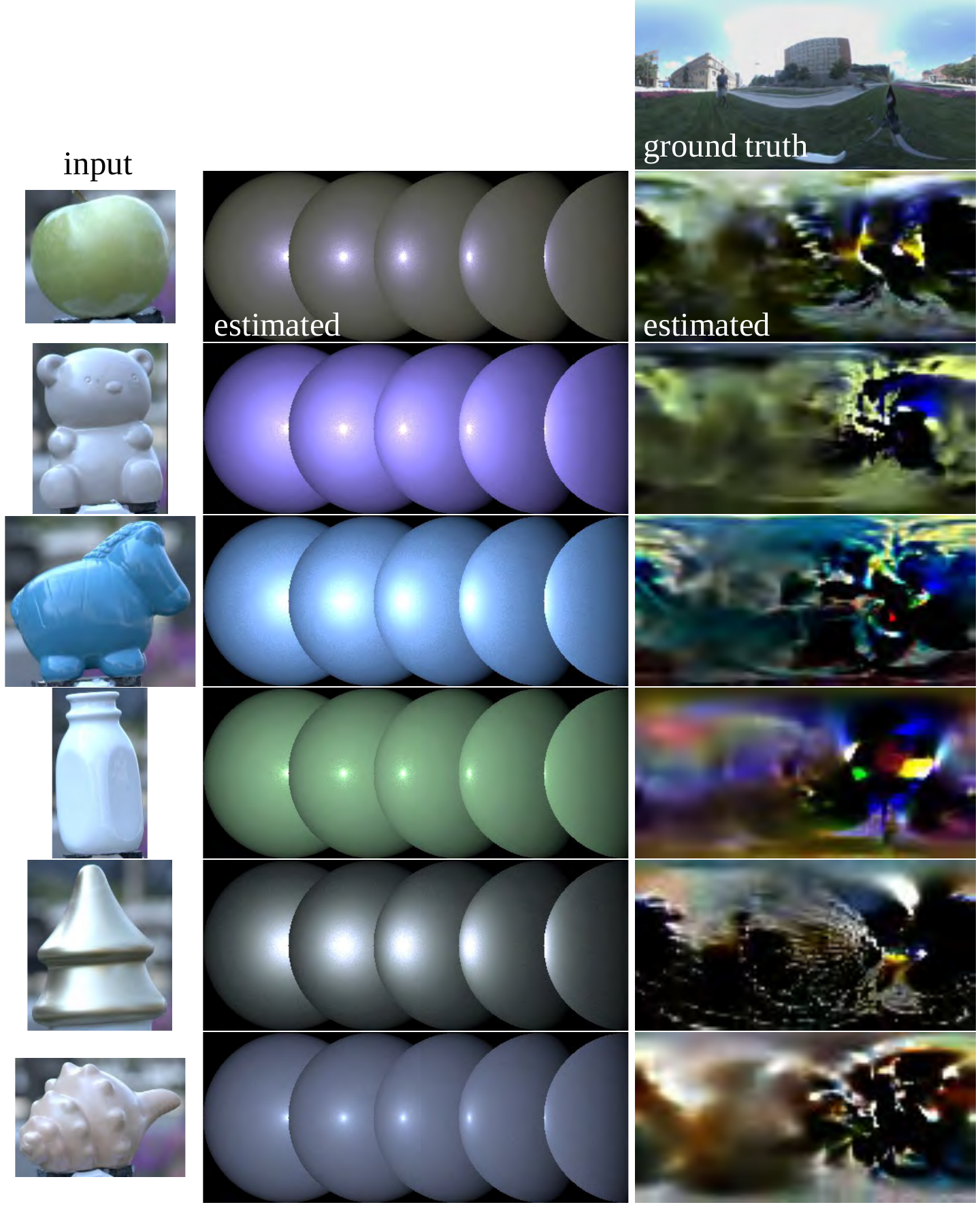}
    \caption{Remaining results of joint estimation of reflectance and illumination from images of real objects [25].}
    \label{fig:full_real} 
\end{figure}

\begin{figure}[t]
    \centering
    \includegraphics[width=\linewidth]{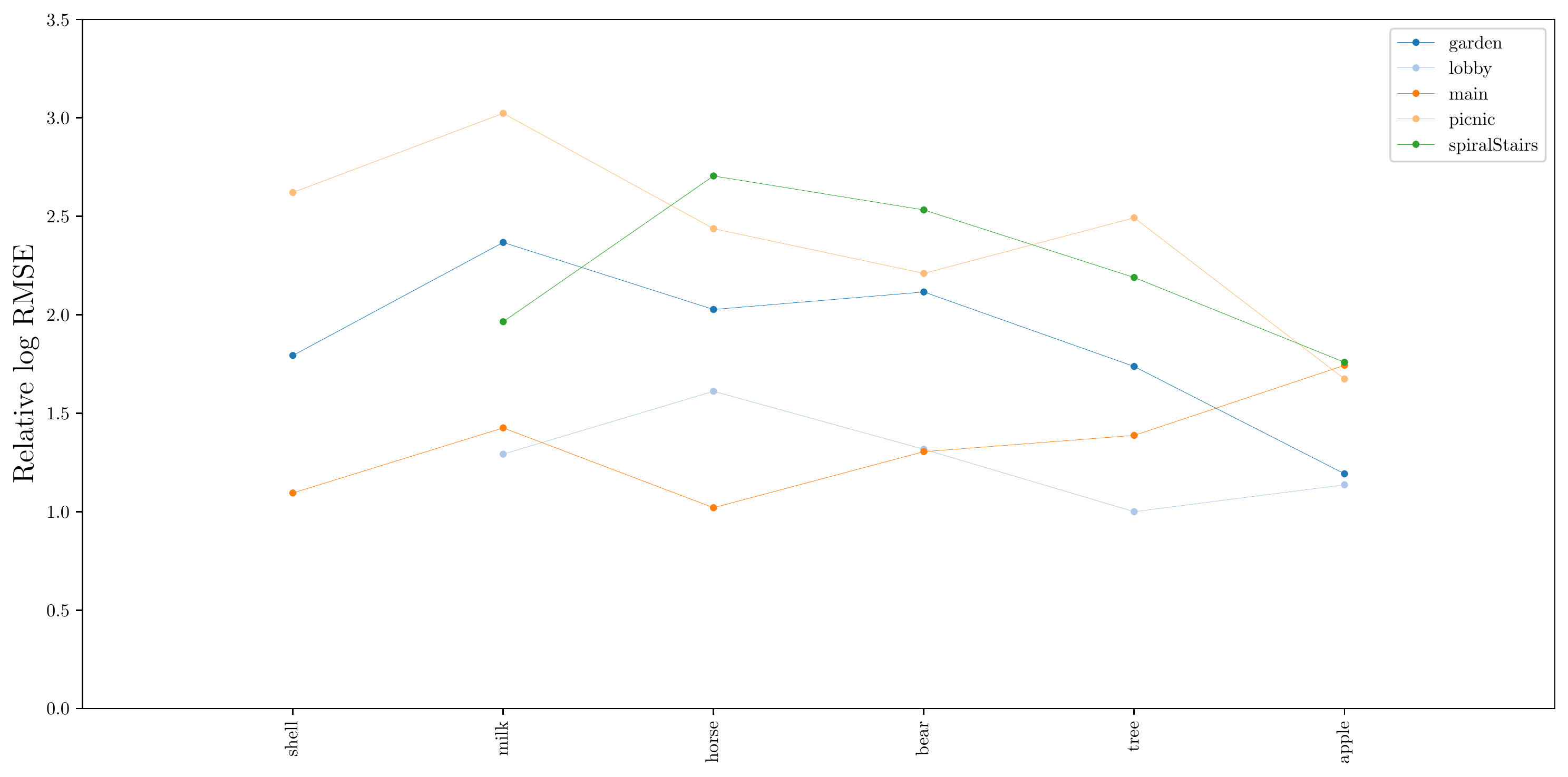}
    \caption{Relative log-space RMSE errors of estimated illumination from joint estimation of Objects Under Natural Illumination Database [25].}
    \label{fig:real_ill_rmse} 
\end{figure}

Finally, we show the remaining results on images of real objects taken under natural illumination from the Objects Under Natural Illumination Database [25]. When combined with Fig. 5(b) in the main manuscript, Fig. \ref{fig:full_real} shows all the results of jointly estimating the BRDF and illumination using input images in the database. The illumination is not as clear as those recovered from synthetic object appearance. This, however, is mainly attributed to the fact that real objects, especially those consisting of flat surfaces like the milk jug, only reflect a portion of the surrounding environment into the camera. Fig. \ref{fig:real_ill_rmse} shows relative log-space RMSE errors of the illumination estimates. The recovered reflectance properties are consistent across results from different illumination, except for the color shifts due to the inherent ambiguity (especially apparent when the object is white). Baking this color constancy problem into the inverse rendering process is left as future work. The errors are larger than the synthetic case, which is also mainly caused by the partial observation captured in the input images. The characteristics of the illumination estimates where the object surface normals partially cover appear consistent across different objects for each environment. Overall, the illumination estimates are quantitatively and qualitatively reasonable, and the BRDF estimates realistic. Note that there are no ground truth measurements for the BRDF, and due to slight errors in geometric calibration of the dataset, direct relighting comparisons were not plausible. These results demonstrate the robustness and accuracy of our method applied to real scenes.

